\documentclass[letterpaper, 10 pt, conference]{ieeeconf}  

\IEEEoverridecommandlockouts                              

\usepackage{amsmath} 
\usepackage{amssymb}  
\usepackage{cite}
\usepackage{amsfonts}
\usepackage{graphicx}
\usepackage{subfigure}
\usepackage[colorlinks,linkcolor=black,anchorcolor=black, citecolor=black, urlcolor=black]{hyperref}
\usepackage{multirow}
\usepackage{color}
\usepackage{gensymb}
\usepackage{algorithm, algorithmic}

\title{\LARGE \bf
Sim-to-Real Grasp Detection with Global-to-Local RGB-D Adaptation
}

\author{Haoxiang Ma$^{*}$, Ran Qin$^{*}$, Modi Shi, Boyang Gao and Di Huang$^{\dag}$
\thanks{This work is partly supported by the National Natural Science Foundation of China (62022011), the Research Program of State Key Laboratory of Software Development Environment (SKLSDE-2023ZX-14), and the Fundamental Research Funds for the Central Universities.}
\thanks{Haoxiang Ma, Ran Qin, Modi Shi and Di Huang are with the State Key Laboratory of Software Development Environment, School of Computer Science and Engineering, Beihang University, Beijing, China.}
\thanks{Boyang Gao is with the Geometry Robotics and the School of Computer Science and Technology, Harbin Institute of Technology, Harbin, China.}
\thanks{* Equal contribution.}
\thanks{\dag Corresponding author. (email:  \href{mailto:dhuang@buaa.edu.cn}{dhuang@buaa.edu.cn}).}
}

\begin{document}

\maketitle
\thispagestyle{empty}
\pagestyle{empty}

\begin{abstract}
This paper focuses on the sim-to-real issue of RGB-D grasp detection and formulates it as a domain adaptation problem. In this case, we present a global-to-local method to address hybrid domain gaps in RGB and depth data and insufficient multi-modal feature alignment. First, a self-supervised rotation pre-training strategy is adopted to deliver robust initialization for RGB and depth networks. We then propose a global-to-local alignment pipeline with individual global domain classifiers for scene features of RGB and depth images as well as a local one specifically working for grasp features in the two modalities. In particular, we propose a grasp prototype adaptation module, which aims to facilitate fine-grained local feature alignment by dynamically updating and matching the grasp prototypes from the simulation and real-world scenarios throughout the training process. Due to such designs, the proposed method substantially reduces the domain shift and thus leads to consistent performance improvements. Extensive experiments are conducted on the GraspNet-Planar benchmark and physical environment, and superior results are achieved which demonstrate the effectiveness of our method. Code is available at \url{https://github.com/mahaoxiang822/GL-MSDA}.
\end{abstract}

\vspace{-3pt}
\section{Introduction}
\vspace{-1pt}

Given its generalizability to new scenes and objects, learning-based grasp detection is being increasingly applied to complex robot manipulation tasks. In this case, a large amount of annotated data is generally required for model training. Unfortunately, it is quite difficult to obtain grasp labels from real-world, which often consumes hundreds of hours in executing grasp candidates to ensure sufficient annotations \cite{DBLP:conf/icra/PintoG16, DBLP:journals/ijrr/LevinePKIQ18}. To deal with this, some attemps \cite{DBLP:conf/iros/DepierreD018, DBLP:journals/ral/ZhangYWZLDZ22, DBLP:conf/icra/EppnerMF21} employ simulators to construct virtual scenes and generate grasp labels, effectively reducing the cost.

Although simulators conveniently enrich the scale and diversity of data, a grasp detection model directly trained with simulated data suffers performance degradation in real-world scenarios. This is due to the discrepancy of data distributions, referred to as the sim-to-real problem. Some studies \cite{DBLP:conf/iros/TobinFRSZA17, DBLP:conf/iros/TobinBDAHKMRSWZ18, DBLP:conf/icra/AlghonaimJ21} introduce Domain Randomization (DR) to alleviate such a performance gap. By randomly varying parameters in the simulator, \emph{e.g.} light direction, object pose and camera perspective, the distribution of simulated data is expected to cover that in the real-world. However, as the true distribution is unknown, it is uncertain that DR can perform data augmentation tailored to the target domain and many uncorrelated data are simultaneously generated during this process, making these solutions less efficient. With the advancements of transfer learning, Domain Adaptation (DA) methods have been investigated in grasp detection \cite{DBLP:conf/icra/BousmalisIWBKKD18, DBLP:conf/icra/FangBHSK18, DBLP:conf/iros/ZhuLBCL0TTL20, DBLP:conf/nips/SaxenaDKN06}. By making use of unlabeled real-world data, they encourage the model to learn features consistent with both simulated and real-world environments. For example, Bousmalis \textit{et al.}\cite{DBLP:conf/icra/BousmalisIWBKKD18} exploit Generative Adversarial Network (GAN) to transfer RGB images rendered by simulators to the style of reality and Fang \textit{et al.} \cite{DBLP:conf/icra/FangBHSK18} design a domain classifier with an adversarial loss to align features between source and target domains. 

\begin{figure}[t]
\setlength{\abovecaptionskip}{0pt}
\centering
\subfigure[]{
\includegraphics[width=0.8\linewidth]
{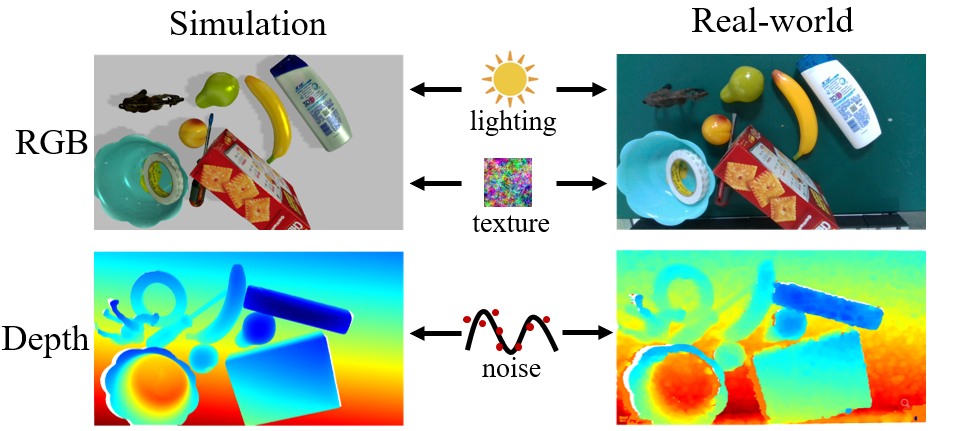}
}

\subfigure[]{ 

\includegraphics[width=0.8\linewidth]{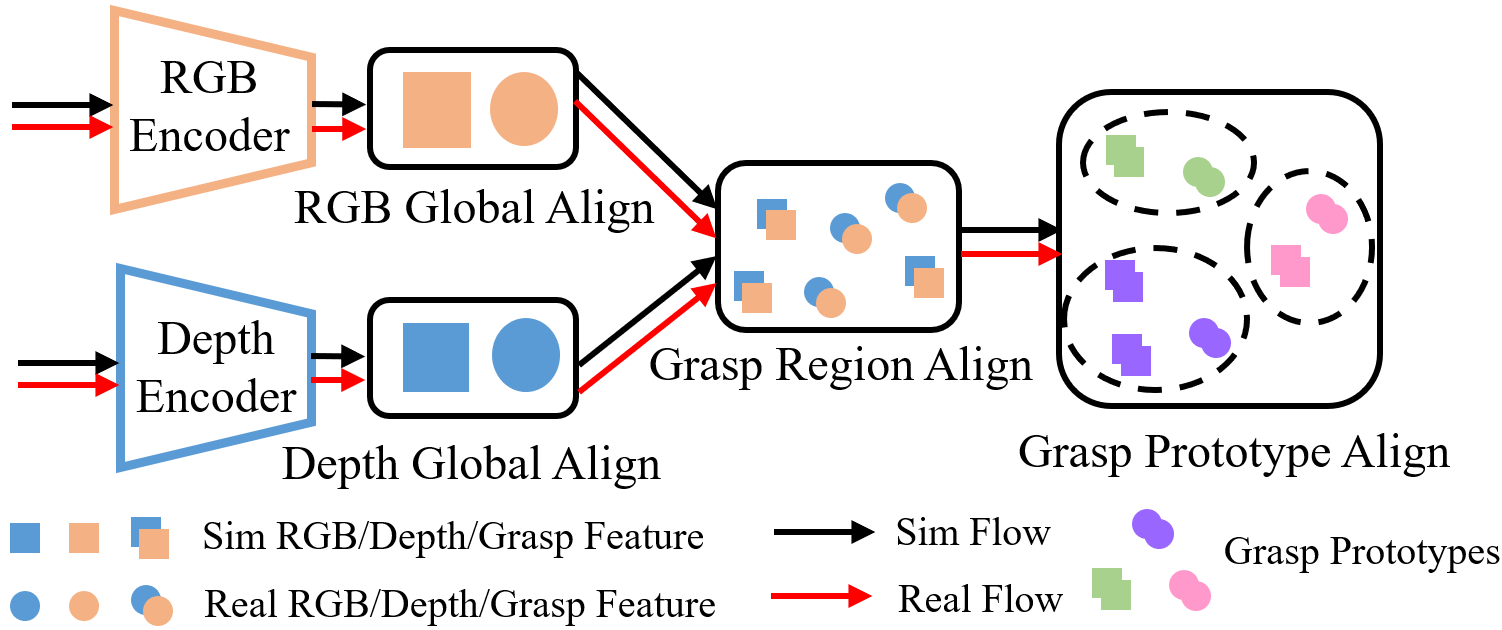}
}
\caption{(a) The domain gap occurs in sim-to-real grasp detection and (b) the proposed GL-MSDA pipeline.}
\label{fig1}
\vspace{-20pt}
\end{figure}

Despite that promising results are reported in sim-to-real grasp detection, existing DA methods still have two limitations. First, in these methods alignment is only performed on the RGB modality \cite{DBLP:conf/icra/BousmalisIWBKKD18, DBLP:conf/icra/FangBHSK18, DBLP:conf/cvpr/JamesWKKIILHB19} while RGB-D sensors have been widely adopted in recent grasp detection systems \cite{DBLP:conf/icra/QinMGH23, DBLP:conf/icra/GouFZXWL21}. As shown in Fig. \ref{fig1} (a), gaps in RGB and depth images between simulated and real-world are distinct, with RGB discrepancies mainly due to lighting and texture and depth disparities arising from noise in depth camera. To the best of our knowledge, alignment on RGB-D data is not well handled, making current DA solutions to multi-modal sim-to-real adaptation problematic. Second, previous studies \cite{DBLP:conf/icra/BousmalisIWBKKD18, DBLP:conf/icra/FangBHSK18, DBLP:conf/cvpr/JamesWKKIILHB19} typically perform global alignment at the image-level, where local features significantly influence the detection performance\cite{DBLP:conf/corl/Ma022, DBLP:journals/ral/ChuXV18, DBLP:conf/cvpr/FangWGL20, DBLP:conf/icra/QinMGH23}. Considering that the distribution of different local shapes varies, directly aligning such features inadvertently results in pulling in grasping features that correspond to entirely dissimilar local shapes, thereby incurring ambiguity.

To address the issues mentioned above, we propose a novel sim-to-real grasp detection framework, namely Global-to-Local Multi-modal Self-supervised Domain Adaptation (GL-MSDA). Fig. \ref{fig1} (b) shows an overview. Specifically, GL-MSDA first introduces self-supervised rotation pre-training to enable two independent networks to learn domain invariant features from simulated and real-world RGB and depth images and then applies global domain classifiers \cite{DBLP:journals/jmlr/GaninUAGLLML16} to separately align the features of simulation and real-world data of each modality. Besides global alignment, GL-MSDA incorporates a local domain classifier to align features of grasp proposals and employs consistency regularization to enforce the  consistency between the results of local and global domain classifiers. Furthermore, to align local geometric features with similar shape distributions from simulation and real-world, inspired by \cite{DBLP:conf/cvpr/Zheng0LW20}, we construct local grasp prototypes by partitioning rotation angles. During training, these prototypes are continuously updated while the prototype distance between the two domains is minimized. Thanks to such designs, GL-MSDA effectively reduces sim-to-real domain shift and delivers decent performance gains. The proposed method is experimentally evaluated on the GraspNet-Planar benchmark and in physical environment with competitive results reported. Additionally, to facilitate future research, we generate a large-scale simulated grasp detection dataset based on GraspNet \cite{DBLP:conf/cvpr/FangWGL20} and GraspNet-Planar \cite{DBLP:conf/icra/QinMGH23} using the PyBullet simulator \cite{coumans2021} and DR techniques.

\vspace{-2pt}
\section{Related Work}


\vspace{-2pt}
\subsection{Sim-to-Real Transfer}
\vspace{-1pt}

The objective of sim-to-real transfer is to narrow the performance gap between simulated and real environments. Over the past decade, it has been extensively studied across various fields, such as robot control \cite{DBLP:conf/icra/PengAZA18}, robot manipulation \cite{DBLP:conf/iros/TobinBDAHKMRSWZ18} and autonomous driving \cite{DBLP:conf/corl/SoXJETAAJ22}. Among the methods in the literature, DR and DA have emerged as two predominant alternatives.

The DR methods\cite{DBLP:conf/iros/TobinFRSZA17, DBLP:conf/iros/TobinBDAHKMRSWZ18,DBLP:conf/icra/AlghonaimJ21,DBLP:conf/cvpr/TremblayPABJATC18, DBLP:journals/trob/HorvathEIHF23} follow the assumption if sufficient diversity is presented in the simulated environment, the generalization ability of the model can be guaranteed for good performance in real-world scenarios. Besides, \cite{DBLP:conf/iros/PashevichSKLS19, DBLP:conf/iros/KleebergerVMTRB20} add Gaussian noise and salt-and-pepper noise to the samples generated by the simulator, replicating the real ones. Given labeled data from the source domain and unlabeled data from the target domain, the DA methods endeavor to map the features of the two domains into a shared domain-agnostic feature space, aiming to reduce the disparity between their feature distributions. Several methods\cite{DBLP:conf/cvpr/Chen0SDG18,DBLP:conf/cvpr/SaitoUHS19,DBLP:conf/cvpr/VSGOSP21, DBLP:conf/cvpr/Zhao022, DBLP:journals/fcsc/TianSPM23, DBLP:journals/fcsc/LiuLFWS23} employ the adversarial training strategy to facilitate detectors in extracting domain-invariant features. Another way to align features is to translate target data (\emph{e.g.} images) into source-like ones using style transfer methods\cite{DBLP:conf/wacv/HsuYTHT0020, DBLP:conf/bmvc/RodriguezM19}. In addition to enhancing the similarity between simulated and real data, some studies\cite{DBLP:conf/cvpr/JamesWKKIILHB19, DBLP:conf/corl/SoXJETAAJ22} map source and target domains a predefined intermediate domain.

\vspace{-2pt}
\subsection{Sim-to-real Transfer for Grasp Detection}
\vspace{-1pt}


Regarding grasp detection, collecting real-world data with annotated grasps is typically expensive and time-consuming. Many studies\cite{DBLP:conf/icra/BousmalisIWBKKD18,DBLP:journals/trob/ZhangCLXQGLLZHHMXS23} conduct robotic arm grasping experiments using physical simulators like PyBullet\cite{coumans2021} and Sapien\cite{DBLP:conf/cvpr/XiangQMXZLLJYWY20}, generating RGB-D images along with corresponding grasping annotations. To bridge the distribution gap between simulated and real data, the DR methods\cite{DBLP:conf/iros/TobinFRSZA17,DBLP:conf/iros/TobinBDAHKMRSWZ18} introduce random variations in visual parameters within the simulation environment, thereby increasing the diversity of training set to enhance model generalization. Others \cite{DBLP:conf/icra/IqbalTCLTCLMB20, DBLP:journals/ral/LiCFCYFLDLH22, DBLP:conf/eccv/DaiZLWDLTW22, DBLP:journals/trob/ZhangCLXQGLLZHHMXS23} employ some high-quality simulators equipped with powerful rendering capabilities to mimic realistic grasp scenarios. On the other side, some methods apply DA to align features derived from the simulated and real-world domain. Fang \textit{et al.}\cite{DBLP:conf/icra/FangBHSK18} employ a domain classifier and an adversarial training strategy to compel the network to learn domain-agnostic features. Bousmalis \textit{et al.}\cite{DBLP:conf/icra/BousmalisIWBKKD18} transfer the style of source domain images to match that of the target domain using GraspGAN. Some studies\cite{DBLP:conf/cvpr/JamesWKKIILHB19, DBLP:conf/iros/ZhuLBCL0TTL20} also make use of artificially defined standard simulation environments or the mean teacher network during training. Although the DA methods achieve a great success, they do not take into consideration the disparity between depth and RGB modalities in the grasping task. Furthermore, they only align the source and target domains but disregards the distribution discrepancy of the features of local grasp regions. In contrast, this paper presents a solution to grasp detection which particularly addresses the sim-to-real problem in a multi-modal RGB-D mode with both global and local feature alignment.

\begin{figure*}[ht]
\vspace{5pt}
\setlength{\abovecaptionskip}{0pt}
\centering

\includegraphics[width=0.7\linewidth]{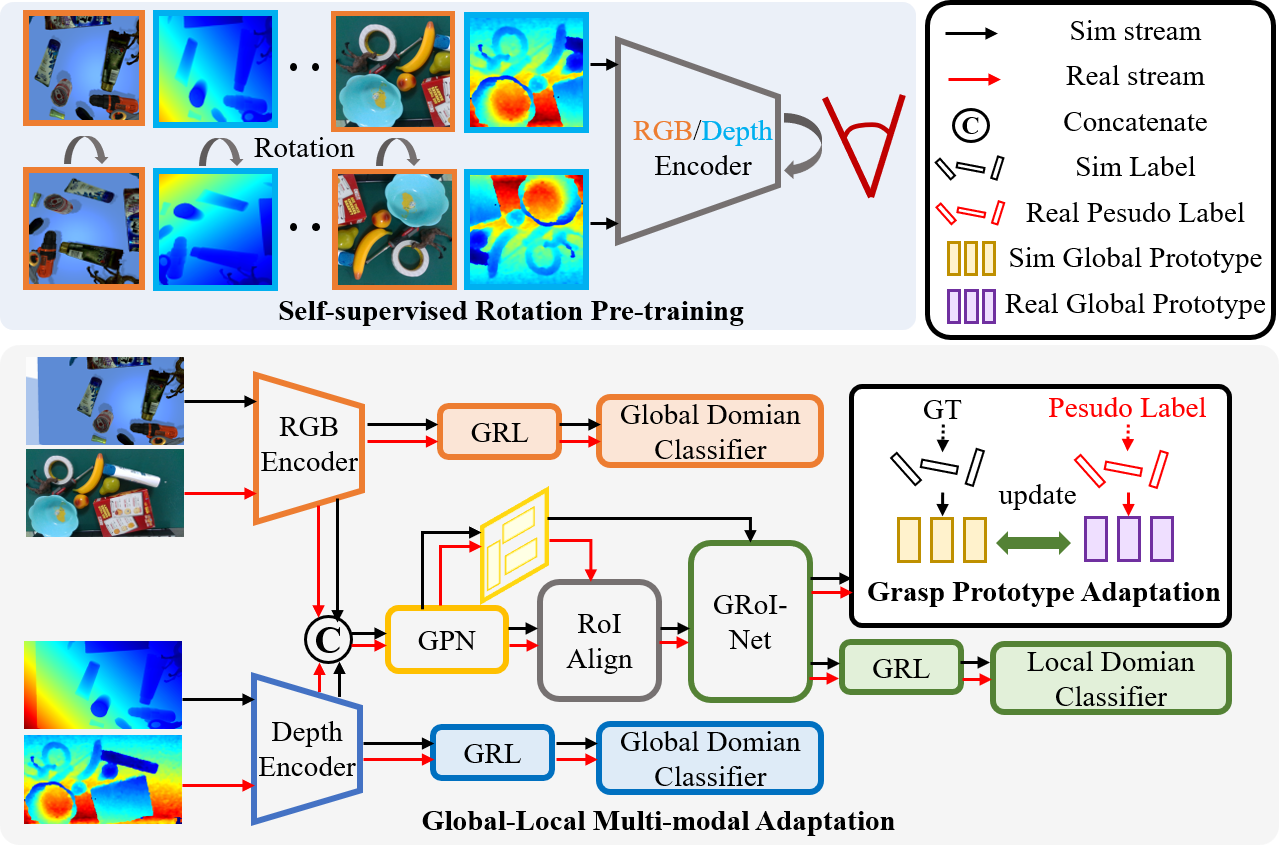}
\label{overall framework}

\caption{Overview of the proposed GL-MSDA method.}
\label{framework}
\vspace{-18pt}
\end{figure*}

\vspace{-2pt}
\section{Methodology}
\vspace{-1pt}

In this section, we introduce the sim-to-real RGB-D grasp detection method proposed in this paper. Specifically, in Section \ref{section:A}, we provide a brief overview of the GL-MSDA framework; in Section \ref{section:B}, we describe the self-supervised rotation pre-training strategy; in Section \ref{section:C}, we present the global-local adaptation module based on multi-modal domain classifiers and consistency regularization; and in Section \ref{section:D}, we finally describe the local grasp prototype adaptation module.

\vspace{-2pt}
\subsection{Overall Framework}
\label{section:A}
\vspace{-2pt}

With labeled simulated RGB-D input, GL-MSDA takes the recently  proposed RGB-D planar grasp detection network with depth prediction \cite{DBLP:conf/icra/QinMGH23} as the baseline model and extends it by incorporating another stream for unlabeled real-world RGB-D data. The pipeline of GL-MSDA is depicted in Fig. \ref{framework}.

The proposed method comprises two stage: \emph{i.e.} Self-supervised Rotation Pre-training and Global-Local Multi-modal Adaptation. In the pre-training stage, RGB and depth images from both the simulation and real-world are mixed. Inspired by \cite{DBLP:journals/ral/LoghmaniRPPCV20}, we rotate the images and employ relative rotation angle prediction as the pretext task for self-supervised training. In the adaptation stage, we train the grasp detection network with labeled simulated data and unlabeled real-world data, initializing it with the pre-trained weights. We utilize the RGB and depth encoders to extract features from the corresponding modalities of the two domains. To fulfill image-level alignment, we introduce two global domain classifiers with Gradient Reversal Layers (GRL)\cite{DBLP:conf/icml/GaninL15}. The RGB and depth features are concatenated and fed into Grasp Proposal Network (GPN) to generate grasp proposals. For local grasp features, we employ the Grasp Region of Interest Network (GRoI-Net) to predict the final grasp parameters and introduce a local domain classifier to align local features from both domains. To achieve fine-grained local alignment, the Grasp Prototype Adaptation (GPA) module utilizes local grasp features to update grasp prototypes, ensuring that the same category of grasp prototypes in the simulated and real-world domains are aligned.

\vspace{-4pt}
\subsection{Self-supervised Rotation Pre-training}
\label{section:B}
To acquire robust visual representations, some methods \cite{DBLP:conf/iclr/GidarisSK18,DBLP:journals/ral/LoghmaniRPPCV20} introduce image rotation as a straightforward yet effective pretext task. To narrow the gap of feature distributions of both the modalities between the simulated and real domains before training the grasp detection network, we separately pre-train the RGB and depth network by predicting relative image rotation angles. We illustrate the pre-train process using the RGB branch as an example, with the depth branch being identical except for its use of depth images as input.

As shown at the top of Fig. \ref{framework}, given an RGB image $I$ either from simulation or real-world and its randomly rotated counterpart, $I^{'}$, the prediction target is the relative rotation angle $A_{I \rightarrow I^{'}}$. $I^{'}$ is obtained by the following transformation:
\begin{equation}
    I^{'} = rot90(I,k)
\end{equation}
\noindent where $rot90$ represents a counterclockwise rotation of the image by $(90 \times k) \degree$, $k \in \{0,1,2,3\}$. To predict $A_{I \rightarrow I^{'}}$, we employ the same RGB encoder to process both $I$ and $I^{'}$, resulting in features $F_{I}$ and $F_{I^{'}}$. These two features are adaptively pooled to a fixed size (\emph{i.e.} $16 \times 16$) and then concatenated. Afterward, the concatenated feature is processed through the convolutional layer $c$ and the fully connected layer $f$, ultimately yielding the relative rotation $A_{I \rightarrow I^{'}}$. The equation is displayed below:
\begin{equation}
    A_{I \rightarrow I^{'}} = f(c(Ada(F_{I})||Ada(F_{I^{'}}))
\end{equation}
Here,  $Ada$ is adaptive pooling and $||$ denotes concatenation. To facilitate the training process, we reframe angle regression as a $k$ value classification problem, optimizing it through the cross-entropy loss function.

\subsection{Global-Local Multi-modal Adaptation}
\label{section:C}

Previous sim-to-real grasp detection methods typically focus on image-level domain adaptation. However, in the field of grasp detection, many studies\cite{DBLP:conf/corl/Ma022, DBLP:journals/ral/ChuXV18, DBLP:conf/cvpr/FangWGL20, DBLP:conf/icra/QinMGH23} point out the importance of features of local regions, which motivate us to design global-local alignment.

The patterns of the distribution gaps between RGB and depth data exhibit differences. Discrepancies in RGB images often arise from lighting and texture variations, while those in depth images primarily result from noise. Learning different domain gaps with a single domain classifier is challenging. To tackle this issue, we employ separate global domain classifiers for RGB and depth features.  This enhances the robustness to distinct distribution shifts encountered in each modality through adversarial learning.

Given the output of RGB and depth encoders, represented as $F_I, F_D \in \mathbb{R}^{C \times H \times W}$, we construct two separate global domain classifiers to predict domain labels, denoted as $P_I, P_D \in \mathbb{R}^{1 \times H \times W}$, for features originating from simulation and real-world, as formulated below:
\begin{equation}
    P_{m} = Sigmoid(c(F_{m})), m \in \left[I,D\right]
\end{equation}
Here, $c$ represents $1\times 1$  convolutional layers. To extract features robust against domain shifts, we establish a min-max game. The objective of the domain classifiers is to accurately classify the origin domain of input features, while the RGB and depth encoder network strive to learn similar features for both simulated and real input, thereby confusing the domain classifiers. Given the domain label $Q$, the loss for the global domain classifier, denoted as $L_I^{DC},L_D^{DC}$ can be formulated as follows:
\vspace{-0.2cm}
\begin{equation}
\begin{split}
    L_{m}^{DC} =-\frac{1}{H \times W} \sum_{i, j} [ Q \log P_{m}^{i, j} + \\
    (1-Q) \log \left(1-P_{m}^{i, j}\right) ], m \in \left[I,D\right]
\end{split}
\end{equation}

To facilitate end-to-end training, we incorporate a GRL between the domain classifier and the encoder network. GRL works by reversing the gradient during the backpropagation phase: while the forward pass remains unchanged, the gradient sign is flipped in the backpropagation. Additionally, we introduce a local domain classifier to align multi-modal grasp features, which assists in reducing disparities within the grasping region, such as variations in object texture and shape. For the $n$th grasp region, denoted by domain label $Q_n$ and prediction $P_n$, the local domain classification loss, denoted as $L_{G}^{DC}$, is formulated as follows:
\vspace{-0.2cm}
\begin{equation}
    L_{G}^{DC} =-\frac{1}{N_G} \sum_n \left[ Q_n \log P_n +(1-Q_n) \log \left(1-P_n\right) \right]
\end{equation}
\noindent where $N_G$ represents the number of grasp regions, and similar to the global classifier, we insert a GRL between the grasp features and the local domain classifier. Furthermore, to improve the robustness of GPN across the simulated and real-world domains, we implement consistency regularization like \cite{DBLP:conf/cvpr/Chen0SDG18} to reinforce the consistency of predictions from the global and local domain classifiers. Specifically, the consistency regularization loss is formulated as follows:
\vspace{-0.2cm}
\begin{equation}
    L_{m}^{CR} = \frac{1}{N_G} \sum_n\left\|\frac{1}{\left|P_{m}\right|} \sum_{i, j} P_{m}^{i, j}-P_n\right\|_2, m\in \left[I,D\right]
\end{equation}

\subsection{Grasp Prototype Adaptation}
\label{section:D}

Local shapes can vary in grasp detection scenarios, which leads to significant differences in the instance-level grasp distribution. Directly aligning local grasp features between simulation and real-world ignores the internal distribution of grasp features, potentially incurring ambiguity by aligning grasp features from different local shapes. \cite{DBLP:conf/cvpr/Zheng0LW20} proposes a prototype-based semantic alignment method to tackle a similar issue in object detection, where the features from the same category are formed into a prototype and the prototype distance in the same category is minimized between the source and target domains. However, for grasp detection, there are no explicit category divisions for constructing prototypes. To address this problem, we introduce a Grasp Prototype Adaptation (GPA) module that generates pseudo-category prototypes for both the simulated and real-world domains based on in-plane grasp rotation angles. GPA aligns these prototypes and iteratively updates them during the training process.

The GPA module aims to minimize the distance between corresponding grasp prototypes from the simulated and real-world domains within the feature space. But two challenges arise: (1) how to divide and construct the grasp prototype; (2)  the prototypes calculated within a small batch may deviate from the true grasp prototypes, rendering them unsuitable for alignment.

For the former issue, we employ the in-plane rotation angle $\theta$ of the local grasp as the division criterion. This choice is made because the distribution of in-plane rotation angles can accurately reflect the local shape distribution in planar grasp detection. We evenly divide the in-plane rotation space into $L$ categories, and based on this division, we construct simulated and real-world grasp prototypes by averaging the features of the grasp regions within each pseudo-category. The equation for this process is as follows:
\vspace{-0.1cm}
\begin{equation}
    P^{S}_{i} = \frac{1}{\left|GT_{i}\right|}\sum_{r \in GT_{i}} F(r)
\end{equation}
\vspace{-0.1cm}
\begin{equation}
    P^{R}_{i} = \frac{1}{\left|GRoI_{i}\right|}\sum_{r \in GRoI_{i}} F(r)
\end{equation}
Here, $P^{S}_{i}$ and $P^{R}_{i}$ represent the $i$th prototype of simulated and real-world domains, $GT_i$ and $GRoI_i$ denote the ground-truth grasp label and the pseudo grasp label predicted by GRoI-Net with the $i$th in-plane rotation angle class and $F(r)$ is the GRoI feature of region $r$. When calculating the simulated prototype, we directly use the ground-truth ($GT$) to extract the grasp region feature and obtain the in-plane angle. For real-world prototype calculation, we rely on the pseudo label predicted by GRoI-Net due to the absence of ground-truth labels. 

For the latter issue, we introduce a weighted moving average to compute the global grasp prototypes, denoted as $GP^{S}$ and $GP^{R}$, using the prototypes $P^{S}$ and $P^{R}$ from each iteration. The update process is defined as follows:
\begin{equation}
    step = \lambda \cdot sim(P^{(t)}_{i},GP^{(t-1)}_{i})
\end{equation}
\vspace{-0,3cm}
\begin{equation}
\label{step_compute}
    GP^{(t)}_{i} = step \cdot P^{(t)}_{i} + (1-step) \cdot GP^{(t-1)}_{i}
\end{equation}
Here, $sim(a,b) = \frac{1}{2} \cdot \left(\frac{a^{T}\cdot b}{\left\Vert a\right\Vert \cdot \left\Vert b\right\Vert}+1\right)$ represents the cosine similarity, and $\lambda$ stands for a small, fixed update step length. We multiply the fixed step length $\lambda$ with the similarity between the prototype and global prototype as the weighted step length. This ensures gradual updates to the global prototypes throughout the training process. To maintain the consistency, we employ the $L_2$ distance to constrain the distance between the prototypes in the simulated and real-world domains, as shown below:
\begin{equation}
    L_{GPA} = \sum_{i \in L} \left\Vert GP_i^S - GP_i^R \right\Vert_2
\end{equation}

\vspace{-0.2cm}
\subsection{Loss Function}
By summarizing the grasp loss $L_{grasp}$ and the losses described above, the overall loss function during training is formulated as:
\begin{equation}
\label{loss_fun}
\begin{split}
    L = L_{grasp} + \alpha \left(L^{DC}_{I} +  L^{DC}_{D}\right) + \beta L^{DC}_{G} \\
    + \gamma \left( L^{CR}_I + L^{CR}_D\right) + \theta L_{GPA}
\end{split}
\end{equation}
$L_{grasp}$ consists of the losses from GRoI-Net and GPN, and for more details, please refer to \cite{DBLP:conf/icra/QinMGH23}.

\begin{table*}[ht]
\small
\setlength{\abovecaptionskip}{0pt}
\setlength{\belowcaptionskip}{0pt}
\vspace{5pt}
\caption{Performance comparison on GraspNet-Planar captured by RealSense.}
\label{comparision with others}
\centering

\begin{tabular}{c|c c c|c c c|c c c}
\hline
\multirow{2}{*}{\textbf{Method}} & \multicolumn{3}{c|}{\textbf{Seen}} & \multicolumn{3}{c|}{\textbf{Similar}} & \multicolumn{3}{c}{\textbf{Novel}}\\
\cline{2-10}
~ & \textbf{AP} & \textbf{AP$_{0.8}$} & \textbf{AP$_{0.4}$} & \textbf{AP} & \textbf{AP$_{0.8}$} & \textbf{AP$_{0.4}$} & \textbf{AP} & \textbf{AP$_{0.8}$} & \textbf{AP$_{0.4}$} \\
\hline
Source Only & 33.93 & 42.28 & 25.08 & 29.68 & 38.25 & 19.09 & 12.00 & 14.79 & 5.03 \\
CycleGAN \cite{DBLP:conf/iccv/ZhuPIE17} & 32.7 & 41.64 & 21.95 & 26.62 & 34.30 & 17.00 & 11.95 & 14.86 & 5.34\\
DAF \cite{DBLP:conf/cvpr/Chen0SDG18} & 37.41 & 46.02 & 28.78 & 33.59 & 41.98 & \textbf{24.31} & 12.83 & 15.96 & 5.65 \\
\hline
GL-MSDA & \textbf{39.25} & \textbf{48.31} & \textbf{30.43} & \textbf{34.38} & \textbf{43.20} & 24.15 & \textbf{13.86} & \textbf{17.34} & \textbf{5.89} \\
\hline
Oracle & 46.45 & 56.15 & 38.32 & 36.23 & 45.34 & 26.83 & 15.43 & 19.39 & 6.44 \\
\hline
\end{tabular}
\vspace{-15pt}
\end{table*}

\section{Simulated Grasp Data Generation}
\label{Data Generation}

\begin{figure}[h]
\setlength{\abovecaptionskip}{0pt}
\vspace{-10pt}
\centering
\subfigure[]{  
\centering  
\includegraphics[width=0.5\linewidth]{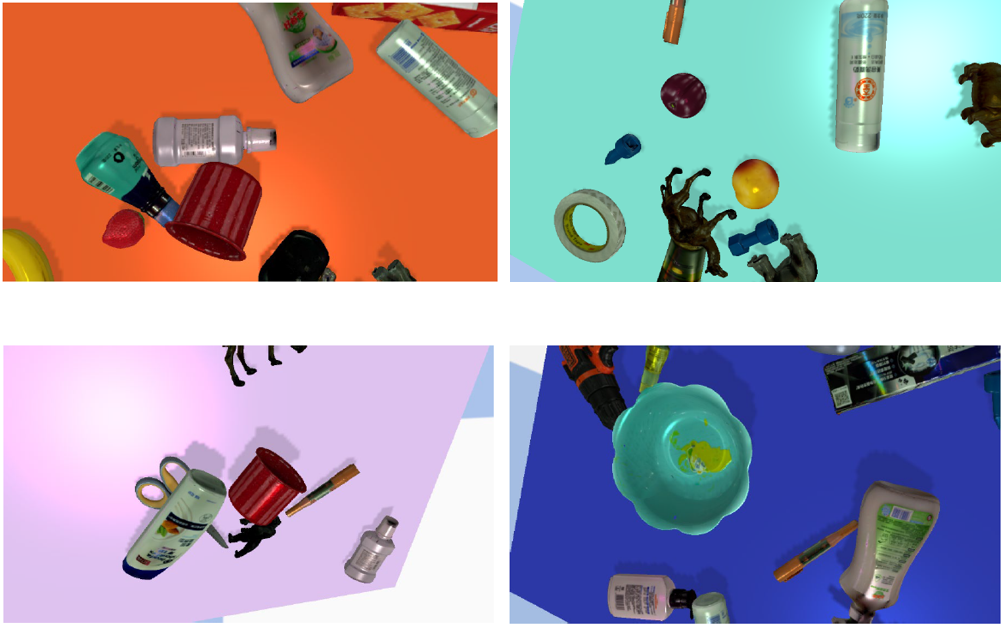}
}
\subfigure[]{ 
\centering    
\includegraphics[width=0.35\linewidth]{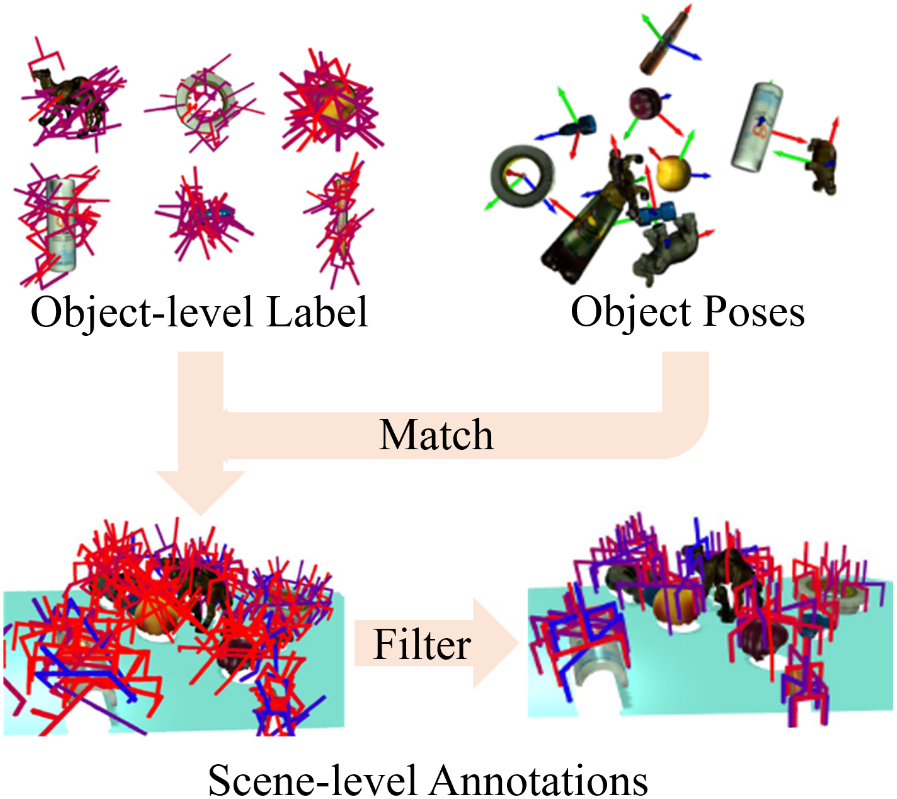}
}
\caption{(a) Visualization of scenes rendered in the simulator by DR. (b) Scene-level grasp annotation.}
\label{simulation}
\end{figure}

In this section, we introduce the simulated grasping dataset used in this paper, namely Sim-GraspNet-Planar. Following \cite{DBLP:conf/icra/QinMGH23}, we use the object set as well as the corresponding grasp annotations in \cite{DBLP:conf/cvpr/FangWGL20} to construct scenes in the PyBullet simulator. To generate a grasping scene, we first set the workspace parameters (\emph{i.e.} desk color and specular reflection) for initialization. Then, we arbitrarily vary the number and category of objects and initialize object poses in the simulator through free fall due to gravity to the desk. Finally, the remaining parameters, such as camera angle and light direction are randomly sampled to capture RGB-D images of the scene from various perspectives. We create 500 scenes, each containing 20 different viewpoints, resulting in a total of 10,000 RGB-D images. Similar to \cite{DBLP:conf/icra/QinMGH23}, camera angles of simulated data are all less than $15\degree$ from the vertical direction of the desktop. Some scenes in our simulation dataset are shown in Fig. \ref{simulation} (a). For grasp annotation, as shown in Fig. \ref{simulation} (b), we employ object-level grasp labels from GraspNet-1billion \cite{DBLP:conf/cvpr/FangWGL20} and project them to scene-level ones with object poses. Gripper poses that collide with either the table or objects, or those whose approach direction is near the table, are filtered out.


\vspace{-0.5cm}
\section{Experiments}

In this section, we evaluate the proposed GL-MSDA method with real-world input on the GraspNet-Planar benchmark \cite{DBLP:conf/icra/QinMGH23} and also in the physical environment.

\subsection{Protocols}

For benchmark evaluation on GraspNet-Planar, we employ Average Precision ($\textbf{AP}_{\mu}$) under different friction coefficient $\mu$ as the metric and the overall result \textbf{AP} is the average of $\textbf{AP}_{\mu}$, where $\mu$ ranges from $0.2$ to $1.0$ with the interval $\Delta\mu=0.2$.

For physical evaluation, we adopt the same setting as in \cite{DBLP:conf/icra/QinMGH23}, where 25 objects of various sizes, shapes and textures from the YCB Object Set \cite{DBLP:journals/ijrr/CalliSBWKSAD17} are used for single object and multi-object grasping. In the single object setting, each object is randomly placed randomly in three different poses and we record the Grasp Success Rate (GSR) as the metric. In the multi-object setting, each cluttered scene is composed of 5 objects and the grasping method works to clean the scene within 10 attempts. Besides GSR, we additionally use Scene Completion Rate (SCR) as the metric.

\vspace{-2.5pt}
\subsection{Implementation Details}
\vspace{-0.5pt}

The simulation data generated in Section \ref{Data Generation} serve as the source domain, while the GraspNet-Planar dataset \cite{DBLP:conf/icra/QinMGH23} is utilized as the target domain. Our network is built upon ResNet-50\cite{DBLP:conf/cvpr/HeZRS16}, with the image dimension set to 1,280 $\times$ 720 pixels. For the ratio of positive and negative samples and NMS threshold, we follow the parameters specified in \cite{DBLP:conf/icra/QinMGH23}. The weights within the loss function in Eq. \ref{loss_fun} are set as follows: $\alpha=0.1$, $\beta=0.1$, $\gamma=0.1$, and $\theta=1000.0$. Additionally, for the fixed update step length in Eq. \ref{step_compute}, we set $\lambda=0.001$. The number of in-plane rotation categories $L$ is set to 12.

Our experiments are launched on 4 RTX 2080 Ti GPUs, with a batch size set to 8 and the allocation of two RGB-D images on each GPU. We apply the SGD optimizer, with the momentum and regularization parameter set to 0.9 and 0.0001, respectively. The learning rate is initially set at 0.005, and then reduced to 0.0005 after 64,000 iterations. We conduct a total of 96,000 training iterations and begin computing $L_{GPA}$ after 56,000 iterations.

\begin{figure}[h]
\setlength{\abovecaptionskip}{0pt}
\vspace{-10pt}
\centering
\subfigure[]{  
\centering  
\includegraphics[width=0.28\linewidth]{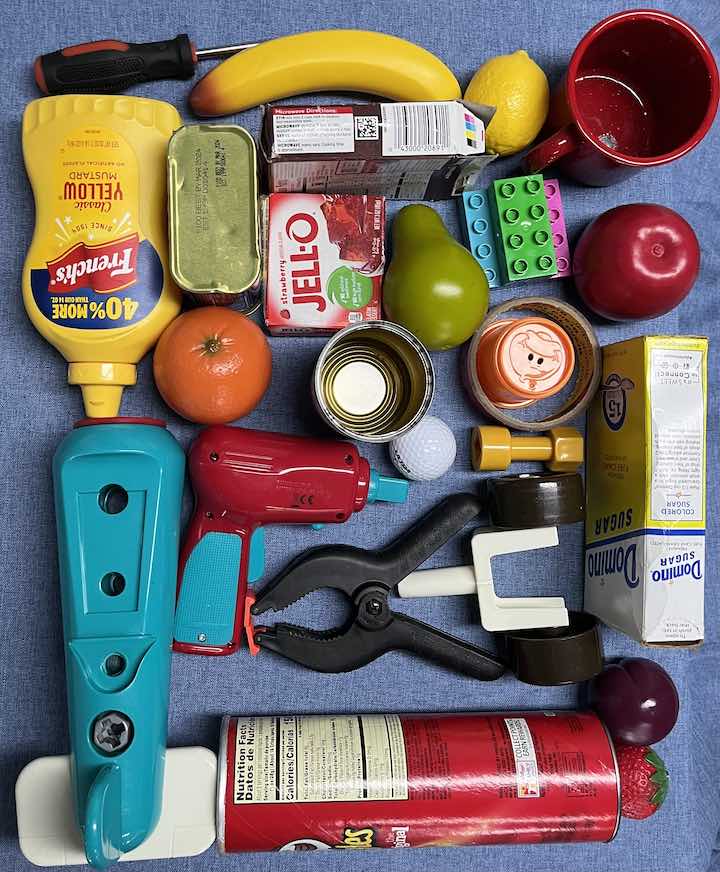}
\label{object set}
}
\subfigure[]{ 
\centering    
\includegraphics[width=0.45\linewidth]{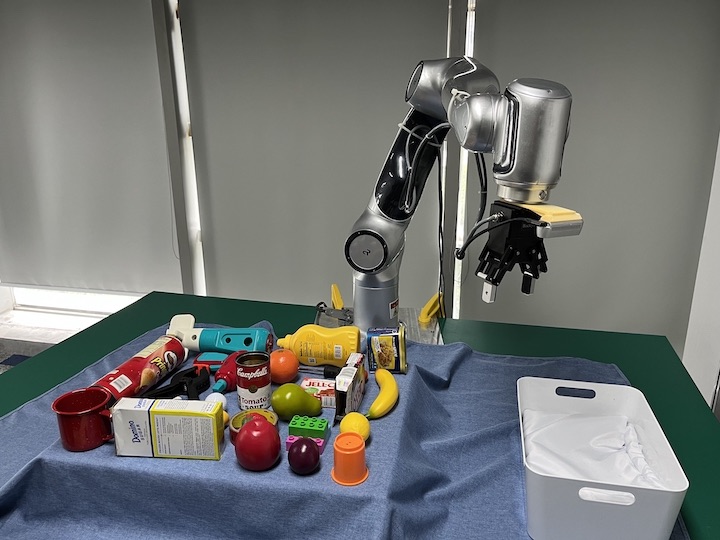}
\label{real-robot setting}
}
\caption{(a) The 25 objects used for physical evaluation. (b)The 7-DoF Agile Diana-7 robot arm with Intel RealSense D435i camera mounted at the end.}
\label{real-robot experiments}
\end{figure}

In physical evaluation, we employ a 7-DoF Agile Diana-7 robot arm, and RGB-D images are captured using an Intel RealSense D435i camera mounted at the end of the arm as shown in Fig. \ref{real-robot experiments} (b). The inference process is executed on a single NVIDIA GeForce 1080 GPU.

\vspace{-0.3cm}
\subsection{Benchmark Evaluation}
\vspace{-0.5pt}

\begin{figure}[h]
\setlength{\abovecaptionskip}{0pt}
\centering
\includegraphics[width=0.8\linewidth]{
{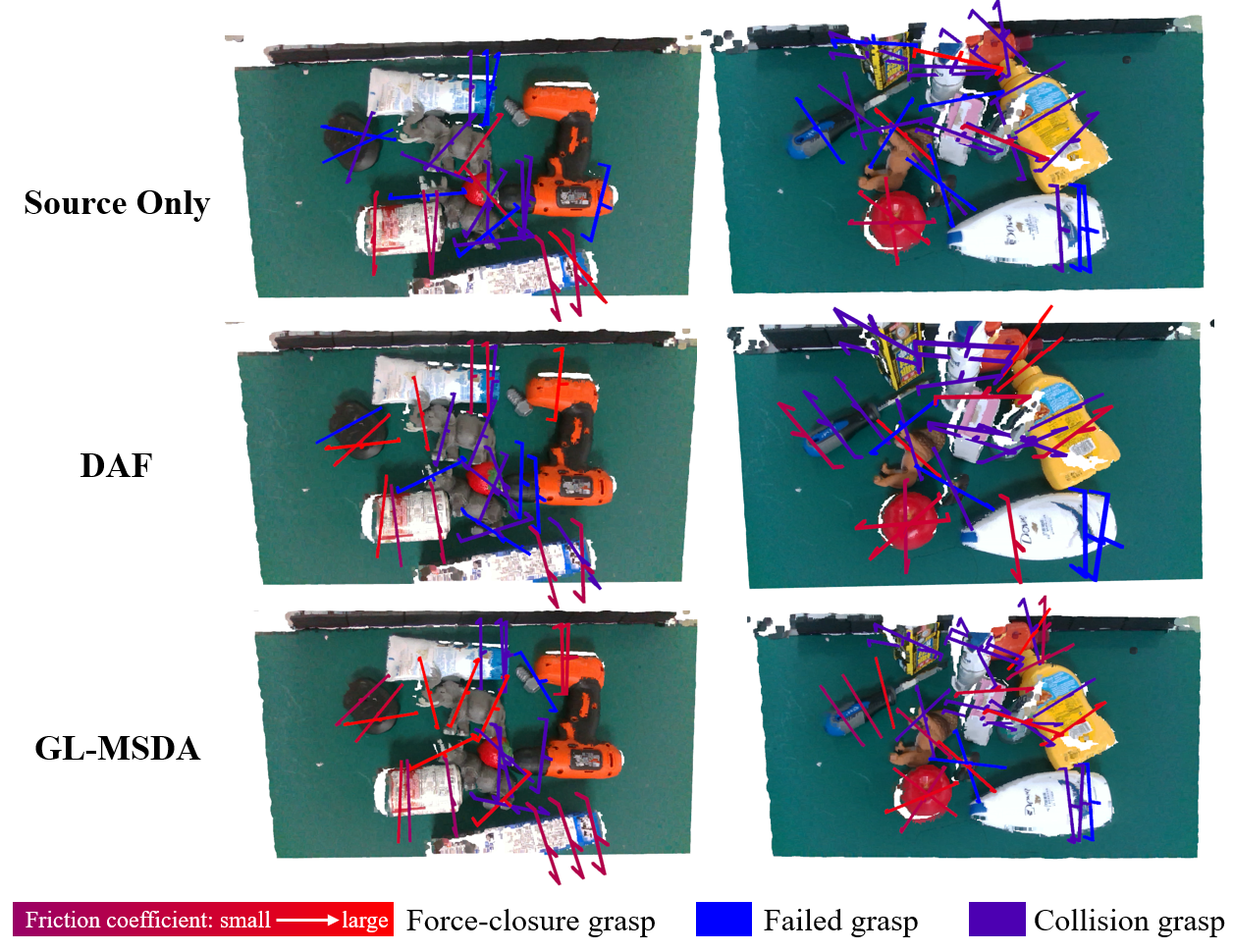}
}
\caption{Visualization of the results on GraspNet-Planar.}
\label{fig4}
\end{figure}

We conduct a comparative evaluation of our method and some representative sim-to-real ones on the GraspNet-Planar benchmark, as shown in Table \ref{comparision with others}. The \textbf{Oracle} model refers to the model trained with labeled real-world data from GraspNet-Planar, which can be seen as the upper bound of the sim-to-real adaptation paradigm. Both the image-level adaptation \cite{DBLP:conf/icra/BousmalisIWBKKD18} and feature-level adaptation \cite{DBLP:conf/icra/FangBHSK18} methods are taken as the counterparts. Due to the differences in the input modalities and network architecture, we replicate the DA methods utilized in the aforementioned two studies on our baseline. Specifically, we adopt DAF \cite{DBLP:conf/cvpr/Chen0SDG18} for feature-level adaptation and CycleGAN \cite{DBLP:conf/iccv/ZhuPIE17} for image-level adaptation. It should be noted that DAF and CycleGAN only differ in the DA module compared to our method and their planar grasp detection networks are consistent with ours for fair comparison. It can be seen that GL-MSDA performs better than its counterparts and compared to the source-only model based on DR, it improves the performance by 4.68\%, 5.01\%, and 2.19\% on seen, similar and novel objects respectively. 
Additional comparison results of different methods are shown in Fig. \ref{fig4}. Compared to the source-only model, our method exhibits a significant advantage. Regarding DAF, thanks to separate adaptation on the depth branch and fine-grained local alignment, our method works better on regions with complex shapes (elephant in the left column) or where accurate prediction of depth is required (screwdriver in the right column).
 
\vspace{-0.5cm}
\subsection{Physical Evaluation}

We also compare our method to the source only model in real-robot experiments. As shown in Table \ref{physical robot experiment}, for single object grasping, our method delivers a large improvement in terms of GSR. For multi-object grasping, our model achieves a gain of 19.83\% in GSR and 16\% in SCR. Furthermore, the results of multi-object scenes are close to the oracle model. These results highlight the effectiveness of our method.

\begin{table}[h]
\vspace{10pt}
\setlength{\abovecaptionskip}{-5pt}
\setlength{\belowcaptionskip}{-0pt}
\caption{Physical robot evaluation results}
\vspace{-10pt}
\label{physical robot experiment}
\begin{center}
\resizebox{\linewidth}{!}{\begin{tabular}{c|c|c|c}
\hline
\multirow{2}{*}{\textbf{Method}} & \textbf{Single-Object} & \multicolumn{2}{c}{\textbf{Multi-Object}}\\
\cline{2-4}
~ & \textbf{GSR (\%)} & \textbf{GSR (\%)} & \textbf{SCR (\%)} \\
\hline
Source Only & 40.00 (30/75) & 43.18 (38/88) & 76.00 (38/50) \\
\hline
GL-MSDA & \textbf{62.67} (47/75) & \textbf{63.01} (46/73) & \textbf{92.00} (46/50)\\ \hline
Oracle & 77.33 (58/75) & 65.75 (48/73) & 96.00 (48/50)\\
\hline
\end{tabular}}
\end{center}
\vspace{-10pt}
\end{table}

\begin{table}[h]
\setlength{\abovecaptionskip}{0pt}
\setlength{\belowcaptionskip}{-0pt}
\caption{Ablation study of grasp depth prediction on GraspNet-Planar captured by RealSense.}
\label{ablation study}
\centering
\begin{tabular}{c|c|c|c}
\hline
{\textbf{Method}} & {\textbf{Seen}} & {\textbf{Similar}} & {\textbf{Novel}}\\
\hline
Pre-train + Global  & 38.50 & 32.86 & 13.86\\
Pre-train + Local  & 38.21 & 33.52 & 13.2\\
\hline
Global + Local & 38.61 & 33.33 & 13.84\\
\hline
Pre-train + Global + Local & 39.02 & 34.09 & \textbf{13.89}\\
\hline

Pre-train + Global + Local + GPA & \textbf{39.25} & \textbf{34.38} & 13.86\\
\hline
\end{tabular}
\vspace{-15pt}
\end{table}

\vspace{-0.1cm}
\subsection{Ablation Study} \label{ablation study section}

\textbf{Influence of global and local domain classifier.} 
In GL-MSDA, we employ both the global and local domain classifiers. To verify their effectiveness, we carry out an ablation study on them separately, as shown in the first and second rows of Table \ref{ablation study}. Without the global or local domain classifier, the results decrease by 0.81\%, 0.57\%, 0.69\%, and 0.52\%, 1.23\%, 0.03\%, on the seen, similar and novel set respectively. This demonstrates that for sim-to-real grasp detection, the global domain classifiers of the RGB and depth networks and the local domain classifier of the grasp features enhance alignment at multiple levels, resulting in better performance.

\textbf{Influence of Self-supervised Rotation Pre-training.}
The self-supervised rotation pre-training scheme for RGB and depth networks helps the model learn robust and domain invariant features before training the grasp detection network. As shown in the third row of Table \ref{ablation study}, the performance drops 0.41\%, 0.76\% and 0.05\% in the seen, similar and novel sets without pre-trained weights, indicating the necessity to the performance.

\textbf{Influence of Grasp Prototype Adaptation.} The GPA module aligns grasp features with similar shape distributions. As shown in the fifth row of Table \ref{ablation study}, it achieves an improvement of 0.23\% on seen objects and 0.29\% on similar objects, demonstrating the effectiveness of fine-grained alignment.

\section{Conclusion}

In this paper, we propose a sim-to-real RGB-D grasp detection method, GL-MSDA. A multi-modal DA framework is designed to enhance the robustness to the domain gaps in RGB and depth modalities. The usage of local adaptation eases domain shift of instance-level grasp features between simulation and real-world. Moreover, we notice the intra-domain distribution of grasp features and present the GPA module which aligns local grasp features more sufficiently. Additionally, a simulation dataset with DR is constructed. The results of benchmark and real-robot experiments show the superiority of our method.

\newpage
\bibliographystyle{IEEEtran}
\bibliography{IEEEabrv, IEEEexample}

\end{document}